\documentclass[10pt,twocolumn,letterpaper]{article}

\usepackage[pagenumbers]{cvpr} 

\usepackage{graphicx}
\usepackage{amsmath}
\usepackage{amssymb}
\usepackage{booktabs}
\usepackage{times}
\usepackage{epsfig}
\usepackage{comment}
\usepackage{url}

\usepackage{colortbl}
\usepackage{arydshln}
\usepackage{multirow}
\usepackage{bm}

\usepackage[pagebackref,breaklinks,colorlinks]{hyperref}

\usepackage[capitalize]{cleveref}
\crefname{section}{Sec.}{Secs.}
\Crefname{section}{Section}{Sections}
\Crefname{table}{Table}{Tables}
\crefname{table}{Tab.}{Tabs.}

\def\cvprPaperID{4286} 
\def\confName{CVPR}
\def\confYear{2022}

\begin{document}

\title{Video Demoiréing with Relation-Based Temporal Consistency}

\author{Peng Dai$^{1}$
\hspace{0.11cm} Xin Yu$^{1}$ 
\hspace{0.11cm} Lan Ma$^{2}\thanks{Corresponding Author}$
\hspace{0.13cm} Baoheng Zhang$^{1}$
\hspace{0.11cm} Jia Li$^{3}$
\hspace{0.11cm} Wenbo Li$^{4}$
\hspace{0.11cm} Jiajun Shen$^{2}$
\hspace{0.11cm} Xiaojuan Qi$^{1}\footnotemark[1]$
\and $^1$The University of Hong Kong \hspace{1cm} $^2$TCL AI Lab\\ $^3$Sun Yat-sen University \hspace{1cm} $^4$The Chinese University of Hong Kong\\
}

\maketitle

\begin{abstract}
Moiré patterns, appearing as color distortions, severely degrade image and video qualities when filming a screen with digital cameras. Considering the increasing demands for capturing videos, we study how to remove such undesirable moiré patterns in videos, namely video demoiréing. To this end, we introduce the first hand-held video demoiréing dataset with a dedicated data collection pipeline to ensure spatial and temporal alignments of captured data. Further, a baseline video demoiréing model with implicit feature space alignment and selective feature aggregation is developed to leverage complementary information from nearby frames to improve frame-level video demoiréing. More importantly, we propose a relation-based temporal consistency loss to encourage the model to learn temporal consistency priors directly from ground-truth reference videos, which facilitates producing temporally consistent predictions and effectively maintains frame-level qualities. Extensive experiments manifest the superiority of our model. Code is available at \url{https://daipengwa.github.io/VDmoire_ProjectPage/}.          

\end{abstract}

\vspace{-0.1in}
\section{Introduction}
\begin{figure}
    \centering
    \includegraphics[width=0.98\linewidth]{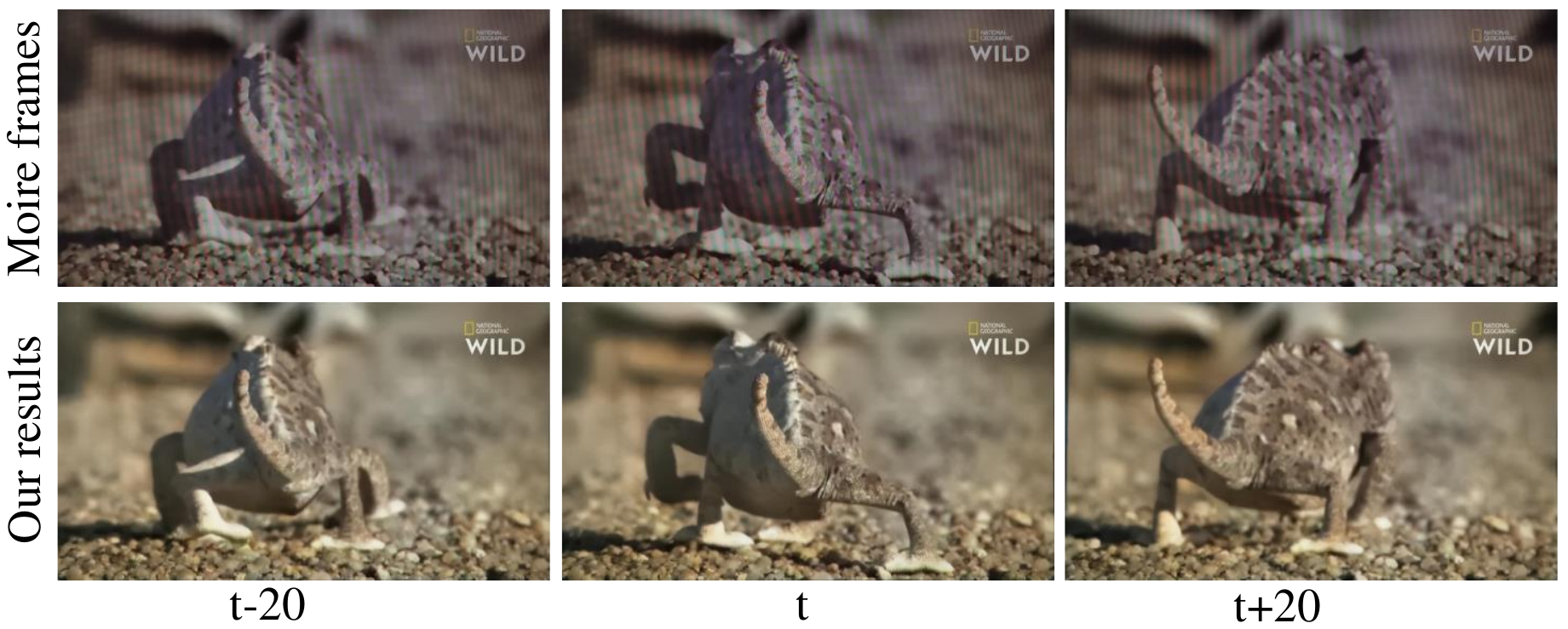}
    \vspace{-0.12in}
    \caption{The first row shows moiré frames at different times, and the second row shows our demoiréd results. Please see our videos, which are clean and temporally consistent.}
    \vspace{-0.22in}
    \label{fig:teaser}
\end{figure}

Video 
is an important source of entertainment, information recording and dissemination through social media.
When photographing a video on a screen, frequency aliasing leads to moiré patterns (Fig.~\ref{fig:teaser}) which appear as  colored stripes, severely degrading the visual quality and fidelity of captured contents. Although many research efforts have been made to remove such moiré patterns in a single image~\cite{he2019mop, sun2018moire, liu2020wavelet, he2020fhde, zheng2020image, nishioka2000endoscope} and attained notable progress with deep learning~\cite{he2019mop, sun2018moire, liu2020wavelet, he2020fhde, zheng2020image}, 
video demoiréing is still an unexplored research problem as far as we know, which is yet of great significance due to the ubiquity and importance of video data in our daily life.

This paper investigates the problem of video demoiréing. 
Compared to image demoiréing, this task offers more opportunities for high-quality frame-level restoration through leveraging auxiliary information from nearby video frames but is yet more challenging as it requires not only frame-level visual quality but also temporal consistency. 

\begin{figure}
    \centering
    \includegraphics[width=0.98\linewidth]{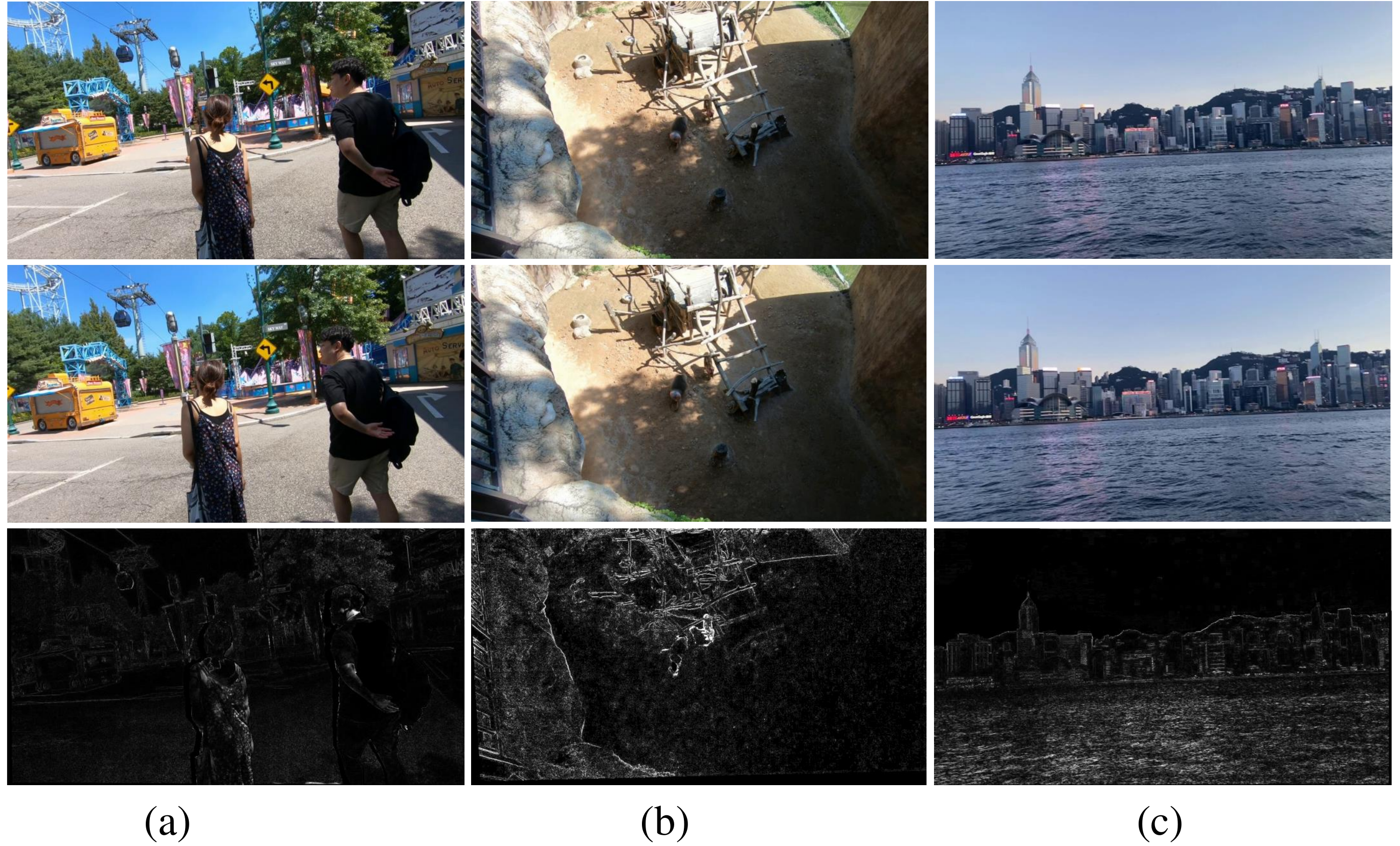}
    \vspace{-0.15in}
    \caption{The problems of flow-based temporal consistency. The first two rows are two consecutive frames, and the last row visualizes the warping error using RAFT~\cite{teed2020raft}. (a) Intensity changes when the person walks from shadow to sunlight. (b), (c) show misalignment between two frames.}
    \vspace{-0.22in}
    \label{fig:flow_tmp}
\end{figure}

The state-of-the-art image demoiréing method~\cite{zheng2020image} fails to recover temporally consistent videos due to its inability to access temporal information/supervision. 
Using existing post-processing methods such as~\cite{lei2020blind, lai2018learning}; in doing so, however, the chance is lost to utilize video information for enhancing frame-level quality.
Besides, these post-processing methods are susceptible to artifacts in demoiréd results, and complicate the system design, leading to increased computational costs. 
Another widely adopted strategy is to incorporate a flow-based consistency regularization~\cite{zhang2021learning, sajjadi2018frame,lei2019fully, zhang2019internal}  on the predicted videos during training, which encourages aligned pixels from nearby frames to have the same pixel intensity values. 
While simple, such regularization ignores natural intensity changes of pixels in videos (Fig.~\ref{fig:flow_tmp} (a)),
is prone to errors in estimated optical flows (Fig.~\ref{fig:flow_tmp} (b) and (c)), and has the potential to propagate artifacts of one frame to nearby frames.
Consequently, the improved temporal consistency tends to sacrifice frame-level quality and fidelity, leading to blurry and low-contrast results ({Fig.~\ref{fig: relation} (a): blurry textures}).
In this work, we present a simple video demoiréing model to leverage multiple video frames and a new relation-based consistency loss to improve video-level temporal consistency without sacrificing frame-level qualities. 
Besides, we construct the first hand-held video demoiréing dataset to facilitate further studies on learning-based approaches.

We analyze the characteristics of moiré patterns in videos and develop a video demoiréing baseline model following~\cite{wang2019edvr, sun2018moire, yang2020learning} with a selective aggregation scheme to adaptively combine aligned features and a pyramid architecture to enlarge the receptive field. 
The baseline model can effectively leverage nearby frames for a better frame-level demoiréing. Deep supervision at different scales is adopted during training to facilitate model optimization. 

Moreover, inspired by the observation that human beings can perceive video flickering~\cite{davenport2007consistency} directly from consecutive frames without using explicitly aligned videos, we propose a simple relation-based temporal consistency loss that encourages the direct relations (\eg, pixel intensity differences) of predicted video frames to follow those of ground-truth frames. In particular, we exploit such relations at multiple levels, including pixel level using pixel intensity differences and patch level using intensity statistics (\eg, mean) changes considering different patch sizes.
Instead of constraining intensities of aligned pixels to be identical, our relation-based regularization directly matches the natural relations and changes of nearby video frames with those of ground-truth videos. This simple design bypasses the aforementioned drawbacks of flow-based consistency regularization and avoids sacrificing frame-level qualities while still being able to enforce the model to learn temporal consistency priors from ground-truth videos. 

Further, as there are no available datasets for developing and evaluating video demoiréing methods, we collect a new video demoiréing dataset with a dedicated pipeline to ensure spatial and temporal alignments between moiré videos and corresponding ground-truth ones.

Finally, extensive experiments on our video demoiréing dataset demonstrate the superior performance of our method. In particular, our method obtains {$22\%$} improvements in terms of LIPIS in comparison with MBCNN~\cite{zheng2020image} and more than $75\%$ of users preferred our results when compared with results without using the multi-scale relation-based consistency loss.

\section{Related Work}
\vspace{-0.05in}
\noindent\textbf{Image Demoiréing.} 
Moiré patterns appear when two similar repetitive patterns interact with each other, and it is frequently observed while capturing images on the screen, which severely degrades image qualities. To remove it, early works have studied spectral models~\cite{sidorov2002suppression} and the sparse matrix decomposition method~\cite{liu2015moire}. However, these methods can only remove certain types of moiré patterns.
 With the rising of deep learning, various convolution neural networks~\cite{he2019mop, he2020fhde, sun2018moire, zheng2020image, liu2020self, liu2020wavelet} have been designed for image demoiréing. 
 Sun et al.~\cite{sun2018moire} built the first large-scale image demoiréing dataset and designed a multi-scale architecture to remove moiré patterns. 
 Further, MopNet~\cite{he2019mop} integrates the characteristics of the moiré pattern into the network and achieves a better result. 
 For high-resolution image demoiréing, He et al.~\cite{he2020fhde} designed a two-stage method to simultaneously remove large moiré patterns and preserve image details. In addition to the above methods which design networks in the image domain, some approaches attempt to address this problem from the perspective of frequency domain~\cite{liu2020wavelet, zheng2020image}. Most recently, Liu et al.~\cite{liu2020self} designed a self-supervised learning method to restore the image only from a pair consisting of one focused moiré-degraded image and one defocused moiré-free image. What differentiates our work from the above research efforts is that we study the new task of video demoiréing with a collected dataset,  which provides new opportunities to improve demoiréing qualities by leveraging temporal information.

\begin{figure}
    \centering
    \includegraphics[width=0.98\linewidth]{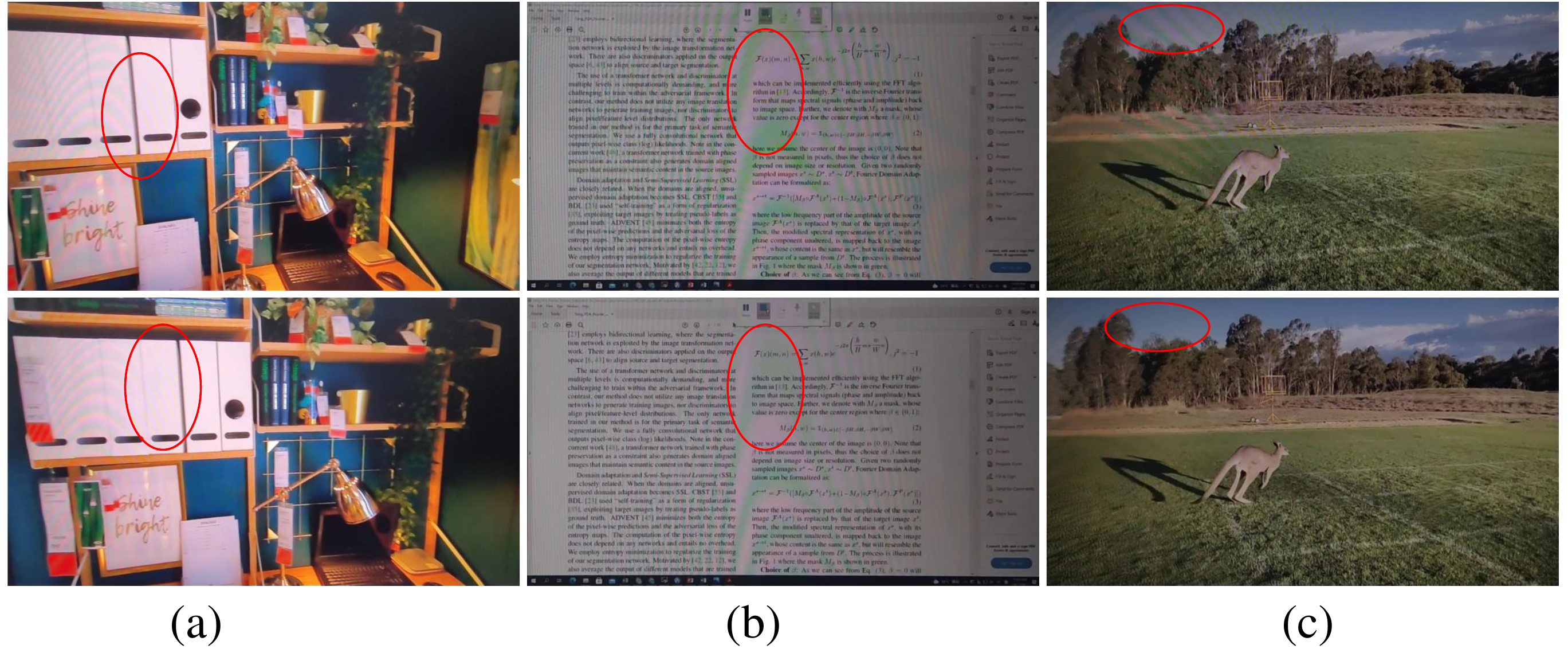}
    \vspace{-0.15in}
    \caption{The characteristics of moiré patterns in the video. Each row represents frames with different time stamps, and the differences between two rows are highlighted by red circles.}
    \vspace{-0.26in}
    \label{fig:moire_char}
\end{figure}

\vspace{0.05in}\noindent\textbf{Multi-Frame Restoration.} 
Multi-frame restoration~\cite{liu2018erase, su2017deep,tassano2020fastdvdnet, tsai1984multiframe, caballero2017real} aims to improve restoration performance by leveraging information from auxiliary frames and typically performs better than image-based counterparts. A key component in multi-frame restorations is the registration of multiple frames, and previous methods usually achieve this using optical flow~\cite{bogoni2000extending, caballero2017real}. Recently, Tian et al.~\cite{tian2020tdan} introduced the deformable convolution~\cite{dai2017deformable} into video super-resolution to implicitly align multiple frames and obtain superior results. This module has been further developed and adopted by several follow-up works~\cite{wang2019edvr, chan2021basicvsr, chan2021basicvsr++, luo2021ebsr}. In this work, we follow the method in ~\cite{wang2019edvr} to align multiple frames in feature space and develop a  module to automatically select valuable information from nearby moiré frames.        


\vspace{0.05in}\noindent\textbf{Video Temporal Consistency.} To obtain temporally consistent videos, previous methods have adopted consistency regularization during network training~\cite{zhang2021learning, sajjadi2018frame, lei2019fully, wang2018video, park2019preserving} or have used it to post-process~\cite{lei2020blind, lai2018learning, bonneel2015blind} flickering videos. 
The most widely adopted consistency regularization is based on dense correspondences (\eg, optical flow), which enforces the intensity of aligned pixels in different frames to be the same~\cite{zhang2021learning, sajjadi2018frame,lei2019fully}. 
However, such a flow-based approach is sensitive to the quality of the estimated dense correspondences~\cite{teed2020raft, dosovitskiy2015flownet} and ignores the natural changes in videos. 
Without optical flows, Lei et al.~\cite{lei2020blind} obtained temporally consistent videos by developing a video prior method which needs time-consuming test-time training. Besides, the effectiveness of the approach relies on a temporally consistent video input which is different from our case.
Some approaches~\cite{ouyang2021internal, wang2020consistent, eilertsen2019single} improve temporal consistency of CNN predictions by augmenting a single frame to multiple frames and enforcing their consistency.
Unfortunately, the moiré pattern in videos is difficult to simulate which makes augmentation-based methods ineffective. Compared to previous works, our relation-based regularization is simple and can take the natural changes of videos into account.
Without using optical flows, our method also avoids suffering from the issues caused by inaccurate optical flow estimation.

\section{Method}
\begin{figure}
    \centering
    \includegraphics[width=0.95\linewidth]{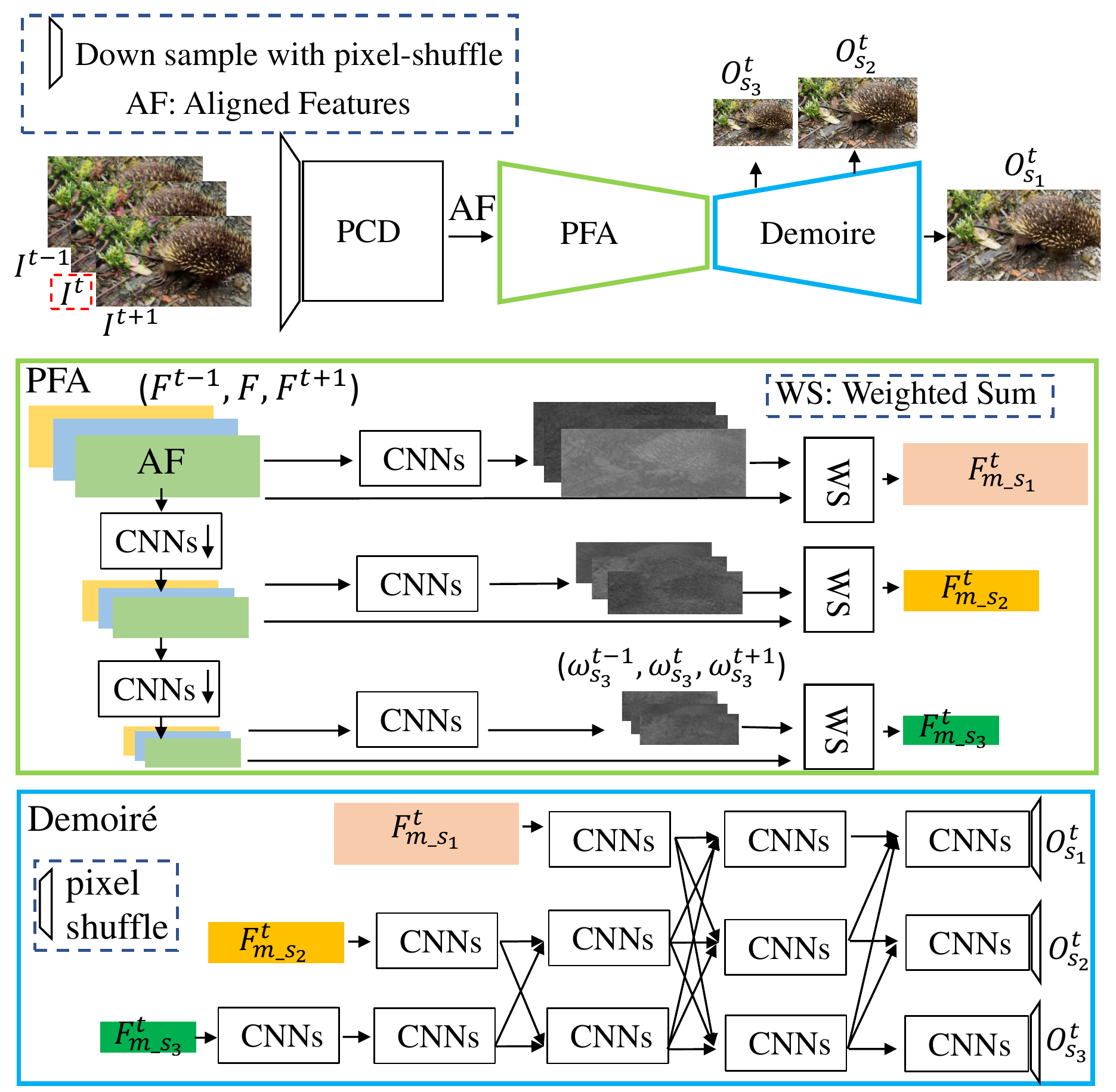}
    \vspace{-0.13in}
    \caption{The overview of our method. Our video demoiréing network mainly consists of three parts: First, the \emph{PCD}~\cite{wang2019edvr} takes consecutive frames as inputs to implicitly align frames in the feature space. Second, the feature aggregation module merges aligned frame features at different scales by predicting blending weights. Third, the merged features are sent to the demoiré model with dense connections to realize moiré artifacts removal.}
    \vspace{-0.24in}
    \label{fig:framework}
\end{figure}

We first present the characteristics of video moiré patterns in Sec.~\ref{section: moire_pattern}, which inspires the design of our baseline video demoiréing model. Then, we elaborate on the key components of our baseline model (Fig.~\ref{fig:framework}) including frame alignment, feature aggregation, and demoiré reconstruction in Sec.~\ref{section: network}. Further, we analyze the weakness of flow-based temporal consistency and detail our newly proposed relation-based consistency regularization in~\ref{section: consistency}. Finally, we show our training objectives in Sec.~\ref{section: loss}.


\subsection{Characteristics of Moiré Patterns in Video}
\label{section: moire_pattern}
The color, shape and location of moiré patterns are generally influenced by camera viewpoints, as shown in Fig.~\ref{fig:moire_char} (a) and (b). 
Under a mild video-capturing setting using hand-held cameras, we observe the following characteristics of moiré patterns in captured videos.
First, as a video plays, the degraded areas have a chance to be clean due to their change of appearing locations (Fig.~\ref{fig:moire_char} (a): the white box at different positions), which can provide valuable information to recover distorted regions in nearby frames. 
Second, the unavoidable hand shaking while shooting videos will slightly change camera viewpoints and induce different moiré patterns in nearby video frames (Fig.~\ref{fig:moire_char} (b): the different text color), which can be leveraged to better distinguish moiré regions by comparing such appearance changes.
Third, the strength of moiré patterns varies in different video frames due to the auto-change of focal length~\cite{liu2020self}, offering a chance to leverage less influenced ``lucky`' frames to restore severely degraded ones (Fig.~\ref{fig:moire_char} (c): the sky with and without moiré patterns). 

Based on the above analysis, our baseline video demoiréing network (Sec.~\ref{section: network}) aligns multiple frames for the purpose of appearance comparisons, effectively aggregates features from nearby frames, and incorporates a blending mechanism to select valuable information from nearby frames in a learnable manner. 

\begin{figure*}
    \centering
    \includegraphics[width=0.95\linewidth]{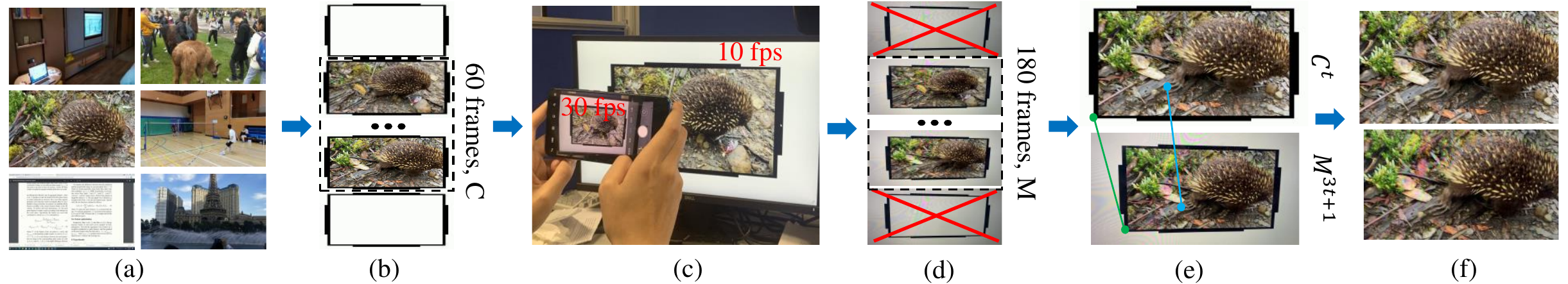}
    \vspace{-0.15in}
    \caption{The pipeline of producing video demoiréing dataset.}
    \vspace{-0.22in}
    \label{fig:data}
\end{figure*}

\subsection{Baseline Video Demoiréing Network} 
\label{section: network}
Our baseline video demoiréing network shown in Fig.~\ref{fig:framework} takes as inputs multiple consecutive video frames {\small $(I^{t-1}, I^t, I^{t+1})$} and outputs restored prediction {\small $O^t$} (equal to {\small $O_{s_{1}}^t$}), leveraging multiple nearby video frames for restoring {\small $I^t$}. Note that we take three adjacent frames to illustrate our model without loss of generality. 

Given the inputs {\small $(I^{t-1}, I^t, I^{t+1})$}, we first incorporate a pyramid cascading deformable (PCD) model in~\cite{luo2021ebsr} to extract and generate implicitly aligned features {\small $(F^{t-1}, F^{t}, F^{t+1})$}. To deal with large moiré patterns in high-resolution videos, we apply pixel shuffle to down-sample the inputs before feeding them into the PCD module which can effectively enlarge the receptive field of the model without sacrificing original information. 

Then, 
 a pyramid feature aggregation (PFA) module (Fig.~\ref{fig:framework}: green box) is developed to selectively aggregate aligned features at multiple scales $(s_1, s_2, s_3)$.
Specifically, the aligned features are down-sampled using convolution layers with a stride of $2$ to produce a feature pyramid that allows feature aggregation to be performed at different resolutions to handle multi-scale moiré patterns.
At each scale $s_i$, the aligned features are concatenated together and used to predict normalized blending weights  ($\omega_{s_i}^{t-1}, \omega_{s_i}^{t}, \omega_{s_i}^{t+1} \in (0, 1)$). The aggregated features $F_{\text{m}\_{s_i}}^t$  are further generated through a pixel-wise weighted summation of aligned features, which enables selective feature aggregation.

Finally, the demoiré reconstruction module 
produces the demoiréd image {\small $O^t$}. 
We densely connect features at different scales to allow them to communicate with each other following~\cite{wang2020deep, yang2020learning} (Fig.~\ref{fig:framework}: blue box).
We apply more convolutional blocks at lower resolution branches to capture a large field of view, benefiting from identifying and removing large moiré patterns and using less convolutional blocks at higher resolution branches to preserve image details.

\subsection{Temporal Consistency}
\label{section: consistency}
Although our baseline video demoiréing network can generate high-quality frame-level results, it cannot ensure video-level consistency. Here, we study the problem of how to generate temporally consistent video demoiréing results.
In the following, we start by analyzing classic flow-based temporal consistency regularization which tends to degrade frame-level qualities, and then elaborate on our simple relation-based temporal consistency loss. 


\vspace{-0.13in}
\paragraph{Flow-Based Temporal Consistency Regularization.}

\begin{figure*}
    \centering
    \includegraphics[width=0.94\linewidth]{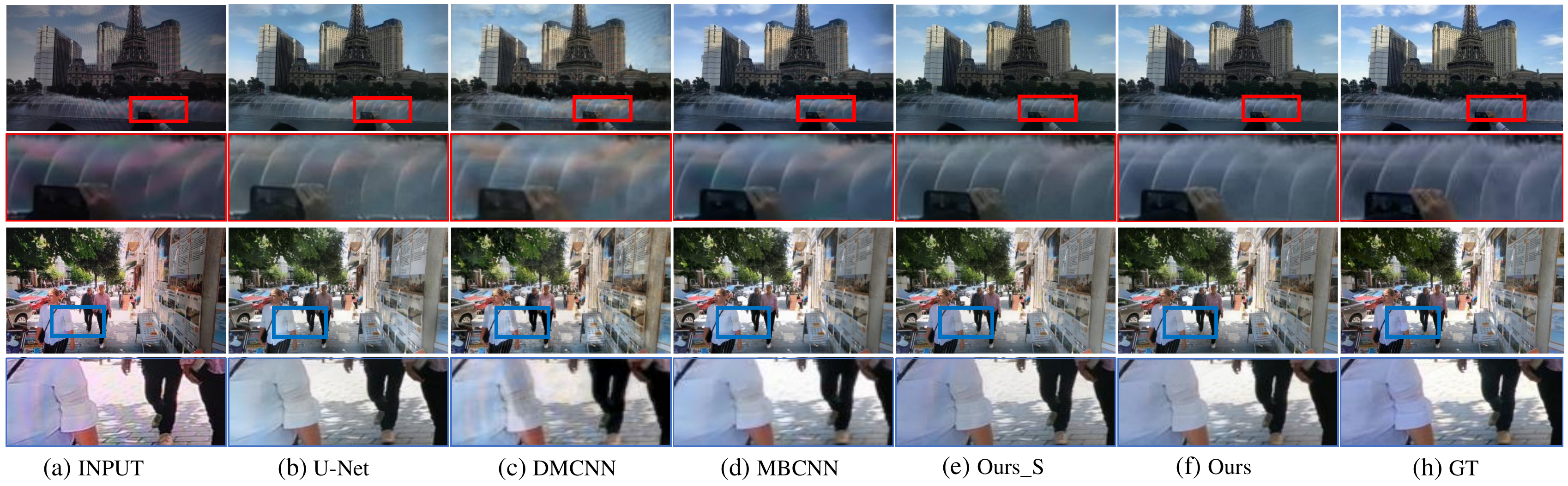}
    \vspace{-0.14in}
    \caption{Qualitative Comparisons. We compare with other baselines and obtain better results on the moiré artifacts removal.}
    \vspace{-0.2in}
    \label{fig: vision}
\end{figure*}

Classic methods achieve temporal consistency by estimating the pixel correspondences in nearby video frames with mostly optical flow methods and building a loss as Eq.~\eqref{equ: flow} to enforce the intensity of matched pixels to be the same~\cite{lai2018learning, zhang2021learning, zhang2019internal}.

\vspace{-0.2in}
\begin{equation}
\vspace{-0.05in}
    \label{equ: flow}
    \begin{small}
    L_{f} = ||M\cdot(\mathcal{W}_{t+1\rightarrow t}(O^{t+1}, \mathcal{F}_{t+1\rightarrow t}) - O^{t})||_{1},
    \end{small}
\end{equation}
where $M$ represents the occlusion map to rule out the influence of occluded pixels, {\small $\mathcal{W}_{t+1\rightarrow t}$} means the flow-based image warp~\cite{ilg2017flownet} to align pixels based on optical flow {\small $\mathcal{F}_{t+1\rightarrow t}$}, and {\small $O^{t}, O^{t+1}$} are nearby output frames. 

\vspace{-0.13in}
\paragraph{Key Observations.} 
We carried out a systematic study on flow-based temporal consistency loss and have the following key observations.
First, a video often undergoes natural changes as time passes due to environmental factors such as lighting and view directions~\cite{pharr2016physically}, and thus a temporally satisfactory video does not necessarily mean that the intensity of the same region never changes (Fig.~\ref{fig:flow_tmp} (a): a person from shadow to sunlight). However, such natural changes will incur a large loss (Fig.~\ref{fig:flow_tmp} (a) third row: the warping error) in flow-based temporal consistency regularization, violating the natural phenomenon. 
Second, the effectiveness of flow-based temporal consistency is adversely affected by the inaccurate estimation of optical flows.  Even the existing state-of-the-art flow estimation method, RAFT~\cite{teed2020raft}, suffers from many failure modes (Fig.~\ref{fig:flow_tmp} (b) and (c): warping errors due to inaccurate flow estimations), especially in objects' boundaries and repetitive textures. These mistakenly matched pixels will incur a penalty that does not exist. 
Finally, the above inaccurate penalties will force the model to trade off frame-level quality for temporal consistency, \eg, averaging matched pixels, leading to blurry and low-contrast results (please see videos and experiments).


\vspace{-0.16in}
\paragraph{Relation-Based Temporal Consistency.} 
Human beings can assess whether a video is temporally consistent or not by directly observing consecutive video frames without using explicitly aligned frames, which motivates us to rethink whether pre-aligned correspondences are needed to learn temporally consistent results and study how to learn temporally consistent results directly from ground-truth reference videos, as they are naturally consistent.
Here, in order to learn temporal consistency patterns from reference videos, we propose matching the direct temporal relations of predicted video frames ({\small $O^{t}, O^{t+1}$}) to those of the reference ones ({\small $G^{t}, G^{t+1}$}), where $G$ indicates the ground-truth video.
The simplest temporal relation can be built by comparing the pixel intensity between video frames; we also investigate other options for temporal relations below.



\vspace{-0.16in}
\paragraph{Basic Relation Loss.}
The most basic relation we consider is the difference between two frames, as Eq.~\eqref{equ: relation}:

\vspace{-0.13in}
\begin{equation}
\vspace{-0.02in}
    \label{equ: relation}
    \begin{small}
    \begin{aligned}
    L_{r} = ||(O^{t+1} - O^{t}) - (G^{t+1} - G^{t})||_{1}.\\
    \end{aligned}
    \end{small}
\end{equation}

As opposed to the flow-based temporal consistency loss in Eq.~\eqref{equ: flow}, which constrains aligned predictions to have the same intensity values, the basic relation loss requires that the difference of outputs and reference frames should be similar, \ie, the predicted results should follow the temporal change of the reference videos.

\vspace{-0.15in}
\paragraph{Multi-Scale Region-Level Relation Loss.} 
Besides pixel-level relations, we also consider region-level relations that follow human habits~\cite{cheng2019review, ng2012human}. {Biologically, the retinal cell receives light from a region instead of a point, and the region size is determined by the distance between retinal cells and observed objects.}
For region-level relations, we use pixel statistics, such as the mean value of pixel intensities, to build the relation loss. We empirically find the mean value works very well in practice. The reason might be that the mean of a patch reflects the brightness of that area, which is closely related to flickers~\cite{choi2018video}.
Specifically, we use patches with different sizes $k\in C$ to take account of various receptive fields, extract the statistics from these patches, and construct a multi-scale region-level relation loss as in Eq.~\eqref{equ: blur_relation}.
Moreover, we only penalize the scale that incurred the minimum difference to protect temporally consistent predictions from nearby potential flickering regions.
\vspace{-0.1in}
\begin{equation}
\begin{small}
\begin{aligned}
    \label{equ: blur_relation}
    &L_{mbr} = \frac{1}{N}\sum_{n=1}^{N}L^{k^*}_{n}|_{k^*=\mathop{\arg\min_{k}\{|(\mathcal{T}_k(O^{t+1}) - \mathcal{T}_k(O^{t}))_n|\}}, k\in C},\\
    &L^{k}_{n} = |((\mathcal{T}_k(O^{t+1}) - \mathcal{T}_k(O^{t}))_n - (\mathcal{T}_k(G^{t+1}) - \mathcal{T}_k(G^{t}))_n|,\\
\end{aligned}
\end{small}
\end{equation}
where $\mathcal{T}_k$ indicates the operation of calculating the statistics of a patch with size $k\in C$ ($C=\{1\}$ is the basic relation-based loss), and $n$ is the pixel position index.

\vspace{-0.15in}
\paragraph{Analysis.} 
The relation-based loss is simple without needing to estimate dense correspondences and thus avoids the problem of misalignment caused by optical flow estimation, and the natural changes in ground-truth videos can be transferred to output frames.
Meanwhile, the model can learn to produce temporally consistent results by mimicking the temporal relations of the reference video, which naturally encompasses temporal consistency priors.

\begin{figure*}
    \centering
    \includegraphics[width=0.94\linewidth]{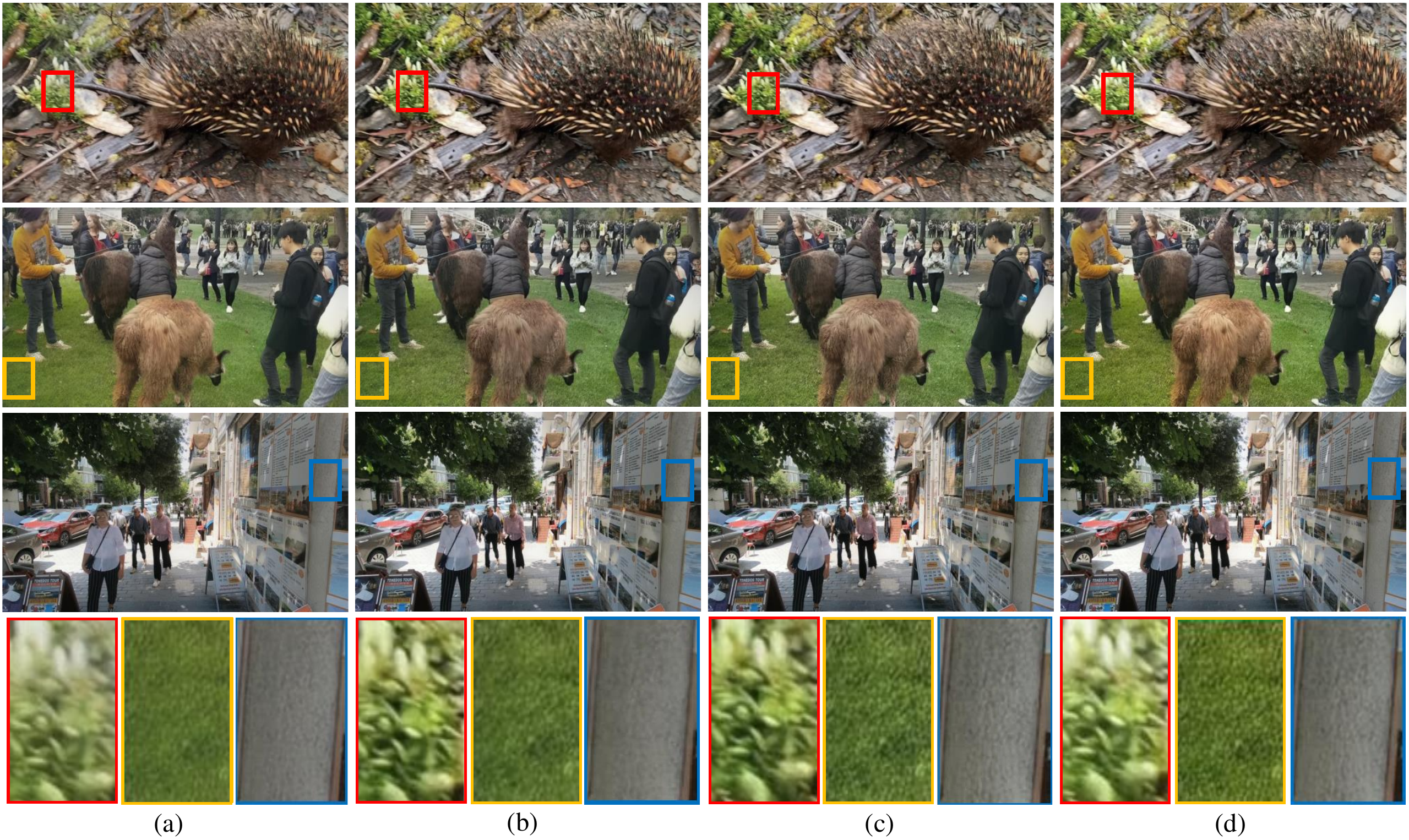}
    \vspace{-0.13in}
    \caption{Different types of temporal consistency. (a) Flow-based temporal consistency. (b) Ours with basic relation loss. (c) The full version of our method. (d) Results without temporal constraints (reference). We can observe that (c) preserves details best.}
    \vspace{-0.2in}
    \label{fig: relation}
\end{figure*}

\subsection{Training Objectives} 
\label{section: loss}
Our overall training objective $L_{train}$, in Eq.~\eqref{equ: train}, is the combination of the frame-level demoiréing loss $L_{d}^{t}$, $L_{d}^{t+1}$,  which regresses outputs at different scales to the ground truths, and the relation loss $L_{mbr}$ of temporal consistency. 
\begin{small}
\begin{equation}
    \label{equ: train}
    L_{train} = L_{d}^{t} + L_{d}^{t+1} + \lambda_{t}L_{mbr},
\end{equation}
\end{small}
$\lambda_{t}$ is used to control the degree of temporal consistency.

To construct $L_d$,  we adopt $L_{1}$ and perceptual loss~\cite{johnson2016perceptual}, which guide the regression process. Apart from the loss on the original resolution, deep supervisions~\cite{lee2015deeply} are applied at different scales to assist the network training. The frame-level demoiréing loss $L_{d}^{t}$ is formulated as Eq.~\eqref{equ: rec}:
\vspace{-0.02in}
\begin{equation}
\vspace{-0.05in}
\label{equ: rec}
\begin{small}
    L_{d}^{t} = \sum_{i, l} || O_{s_i}^{t} - G_{s_i}^{t}||_{1} + \lambda  ||\Phi_{l}(O_{s_i}^{t}) - \Phi_{l}(G_{s_i}^{t})||_{1},
\end{small}
\end{equation}
where $O_{s_i}^{t}$ and $G_{s_i}^{t}$ are output and corresponding ground truth at the $s_i$ scale, respectively. $\Phi_{l}$ is a set of VGG-16 layers, and $\lambda$ is the weight used to balance different parts.

\section{Video Demoiréing Dataset}
We collect the first video demoiréing dataset captured by hand-held cameras, \eg, a smartphone camera.
The capturing pipeline to ensure spatial and temporal alignments between camera-recorded and original videos is shown in Fig.~\ref{fig:data} and elaborated below.

First, the 720p high-quality source videos displayed on the screen  consist of videos from REDS~\cite{nah2019ntire}, MOCA~\cite{lamdouar2020betrayed}, and videos taken by ourselves.  To ensure the diversity of collected videos, we manually choose videos covering various scenarios, including human beings, landscapes, texts, sports, and animals (examples in Fig.~\ref{fig:data} (a)). We collect 290 videos, and each video has 60 frames. 

Second, it is difficult to align videos recorded by cameras and source videos played on the screen considering different frame rates and asynchronous start timestamps. For example, if the camera frame rate is not divisible by the video frame rate, the recorded frame will contain multi-frame information (occurs when switching frames) from the source video, which results in blurry images. Even though the frame rate meets the requirement, different start timestamps (\ie, start to play and record the video) also cause the problem of multi-frame confusion. For these obstacles, we adjust the frame rates and insert start/end flags into videos. Specifically, we set camera and source video frame rates to 30 fps and 10 fps, respectively, and extend source videos with a few white frames at the beginning and the end of each video. What's more, we follow the data collection process in~\cite{sun2018moire} to add some black blocks surrounding the frame to provide more robust keypoints (Fig.~\ref{fig:data} (b) and (c)).

Third, given the source video, mobile phone, and monitor, the moiré pattern can be produced by adjusting the camera view points. While capturing, the mobile phone is hand-held by a person to simulate practical video recording senarios, and different shooting angles and distances are adopted to increase the diversity of moiré patterns (Fig.~\ref{fig:data} (c)). After recording, we can obtain 180 frames (three times the source video) from each video after removing the pre-inserted white frames (Fig.~\ref{fig:data} (d)), and the final moiré frame is sampled among three consecutive frames. Here, we sample the intermediate one since it is not sensitive to frame transitions (Fig.~\ref{fig:data} (e)).

Finally, to obtain training pairs (Fig.~\ref{fig:data} (f)), source and captured frames should be aligned through frame correspondences, such as optical flow and homography matrix. In this work, we adopt the homography matrix to align two frames (Fig.~\ref{fig:data} (e)). Instead of using only keypoints (ORB~\cite{rublee2011orb}) detected on image regions~\cite{he2020fhde} or auxiliary black regions~\cite{sun2018moire}, we utilize both of them to estimate the homography matrix using the RANSAC~\cite{vedaldi2010vlfeat} algorithm. 

\vspace{-0.1in}
\section{Experiments}
In this section, we first introduce training details (Sec.~\ref{sec: training}), then qualitatively and quantitatively compare our method with other baselines at the frame level (Sec.~\ref{sec: frame_level}) and the video level (Sec.~\ref{sec: video_level}). Finally, we validate our video demoiréing model and the relation-based consistency regularization (Sec.~\ref{sec: ablation}).

\subsection{Training Details}
\label{sec: training}
 The video demoiréing network takes three consecutive frames as inputs to predict one restored image. To train the model, we automatically divide the video demoiréing dataset into 247 train videos and 43 test videos, and the hyperparameters $\lambda$ and $\lambda_{t}$ are set to 0.5 and 50, respectively.
 Furthermore, we adopte four region sizes $C=\{1, 3, 5, 7\}$ to simulate different receptive fields. The optimizer in our implementation is Adam with a cosine learning rate~\cite{loshchilov2016sgdr}. In total, we train 60 epochs with batch size 1 on one NVIDIA 2080Ti GPU, and the temporal consistency loss is invoked in the last 10 epochs for training stability.

\subsection{Frame-Level Comparisons}
\label{sec: frame_level}
\begin{figure}
    \centering
    \includegraphics[width=0.93\linewidth]{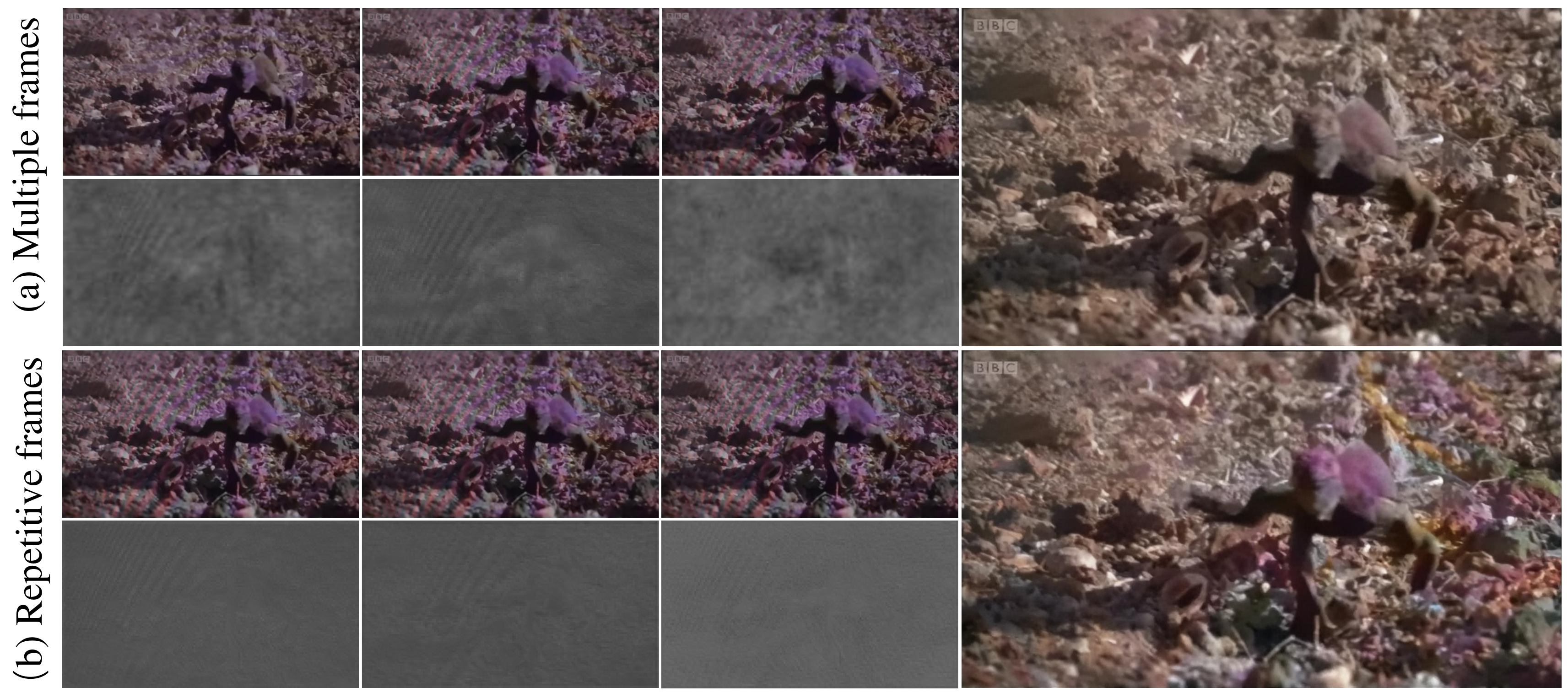}
    \vspace{-0.13in}
    \caption{Visualization of weight maps. (a) Three consecutive frames and the weight maps. (b) Replace consecutive frames with repetitions of a single frame and the weight maps.}
    \vspace{-0.12in}
    \label{fig:weights}
\end{figure}

We compare our approach with image demoiréing methods (\ie, MBCNN~\cite{zheng2020image} and DMCNN~\cite{sun2018moire}) and other widely used backbones, such as U-Net~\cite{ronneberger2015u}. In order to verify the effectiveness of video demoiréing without being affected by other factors (\eg, number of parameters and the choice of loss function), we adopt our video demoiréing model but change the input to repetitions of a single frame (Ours\_S, see Fig.~\ref{fig:weights} (b)). To quantitatively measure the performance of demoiréing, we adopt PSNR, SSIM, and LPIPS~\cite{zhang2018unreasonable} that is more aligned with human perception as our metrics. ('$\uparrow$': larger value is better, '$\downarrow$': smaller value is better.)

 \vspace{-0.05in}
 \begin{table}[!htb]
    \centering
    \resizebox{.7\columnwidth}{!}{
    \begin{tabular}{c|c|c|c}
    \hline
     Methods & LPIPS $\downarrow$ & PSNR $\uparrow$ & SSIM $\uparrow$\\
     \hline
     MBCNN~\cite{zheng2020image} & 0.260 & 21.534 & {\color{red} 0.740}\\
     DMCNN~\cite{sun2018moire} & 0.321 & 20.321 & 0.703\\
     U-Net~\cite{ronneberger2015u} & 0.225 & 20.348 & 0.720\\
    \hdashline
     Ours\_S & {\color{blue} 0.212} & {\color{red} 21.772} & 0.729\\
     Ours & {\color{red} 0.202} & {\color{blue} 21.725} & {\color{blue} 0.733}\\
     \hline
    \end{tabular}}
    \vspace{-0.1in}
    \caption{Demoiréing performance of different methods. ({\color{red} Red: }{best},  {\color{blue} Blue: }{second best})}
    \label{tab: metrcis}
    \vspace{-0.26in}
\end{table}

\begin{table}[!htb]
    \centering
    \resizebox{.9\columnwidth}{!}{
    \begin{tabular}{c|c|c|c|c}
    \hline
     Methods & FID $\downarrow$ & warping error $\downarrow$ & user study $\uparrow$ & \textcolor[rgb]{0.5,0.5,0.5}{LPIPS$\downarrow$}\\
     \hline
     Ours\_S & 0.094 & 5.98 & 14\% & \textcolor[rgb]{0.5,0.5,0.5}{0.212}\\
     Ours & \textbf{0.084} & 5.65 & 25\% & \textcolor[rgb]{0.5,0.5,0.5}{0.202}\\ 
   \hdashline 
     Ours+F & 0.109 & \textbf{2.70} & 9\% & \textcolor[rgb]{0.5,0.5,0.5}{0.339}\\ 
     Ours+R & 0.088 & 4.79 & 42\% & \textcolor[rgb]{0.5,0.5,0.5}{0.211}\\
     Ours+M & 0.085 & 5.03 & - & \textcolor[rgb]{0.5,0.5,0.5}{0.201}\\
     \hdashline
     GT & 0.000 & 4.56 & - & \textcolor[rgb]{0.5,0.5,0.5}{0.000} \\
     \hline
    \end{tabular}}
    \vspace{-0.1in}
    \caption{Temporal consistency measurements when $\lambda_{t}$ is 50. {Ours\_S: video demoiréing model with three repetitive frames, Ours: video demoiréing model with multiple frames, Ours +F: add flow-based consistency loss, Ours+R: add basic relation-based consistency loss, Ours+M: add multi-scale relation-based consistency loss.}
    In user study, all other baselines are compared with Ours+M, and this table reports the percentage of each baseline being selected (Ours+M outperforms all baselines).}
    \label{tab: consistency}
    \vspace{-0.15in}
\end{table}

\vspace{-0.15in}
\paragraph{Qualitative Comparison.}
 In Fig.~\ref{fig: vision}, we show images restored by different methods. It clearly shows that our approach has advantages over other methods for removing moiré artifacts, such as the moiré patterns on the fountain, white T-shirt and floor. We attribute the superiority of our method to its ability to utilize auxiliary information from the nearby video frames.   

\vspace{-0.15in}
\paragraph{Quantitative Comparison.}
 Frame-level quantitative results are reported in Table~\ref{tab: metrcis}. Under the circumstance of single image demoiréing, our method (Ours\_S) outperforms previous methods (above the dotted line). Moreover, the performance is further improved using multiple frames (Ours), especially LPIPS, which manifests the effectiveness in leveraging multiple frames to improve perception results. 
 

\subsection{Video-Level Comparisons}
\label{sec: video_level}
Following previous works~\cite{wang2018video, chang2019free}, we adopt FID and warping error to measure video-level performance. Here, FID measures the distance between output and ground-truth videos in the feature domain using I3D~\cite{carreira2017quo}, and the warping error calculates differences between two frames aligned by optical flows~\cite{teed2020raft}. Note that the warping error cannot accurately reflect the video temporal consistency due to inaccurate optical flow and natural changes in videos. To illustrate it, we calculate the warping error of ground-truth videos (Table~\ref{tab: consistency}: last row), which is still very large. Besides, we also conduct user studies to assist video-level comparisons. 
For the user study, participants are asked to choose one out of two videos based on video quality or mark them as indistinguishable; they are given sufficient time to make the decision. In the process of our user study, two videos produced by different methods are displayed in random order, and participants can replay videos with various frame rates. In total, 14 individuals participated in our experiments.

{As our baseline video demoiréing model (Ours) obtains better results than other compared methods, we take it as the  baseline model for video-level evaluation. Specifically, we compare the video temporal consistency and quality with the results of single image demoiréing (Ours\_S), classic flow-based consistency regularization (Ours+F, replace $L_{mbr}$ loss with $L_f$ loss in Eq.~\eqref{equ: flow}) and multi-scale relation-based consistency regularization (Ours+M, $L_{mbr}$ loss).}

\begin{figure}
    \centering
    \includegraphics[width=0.93\linewidth]{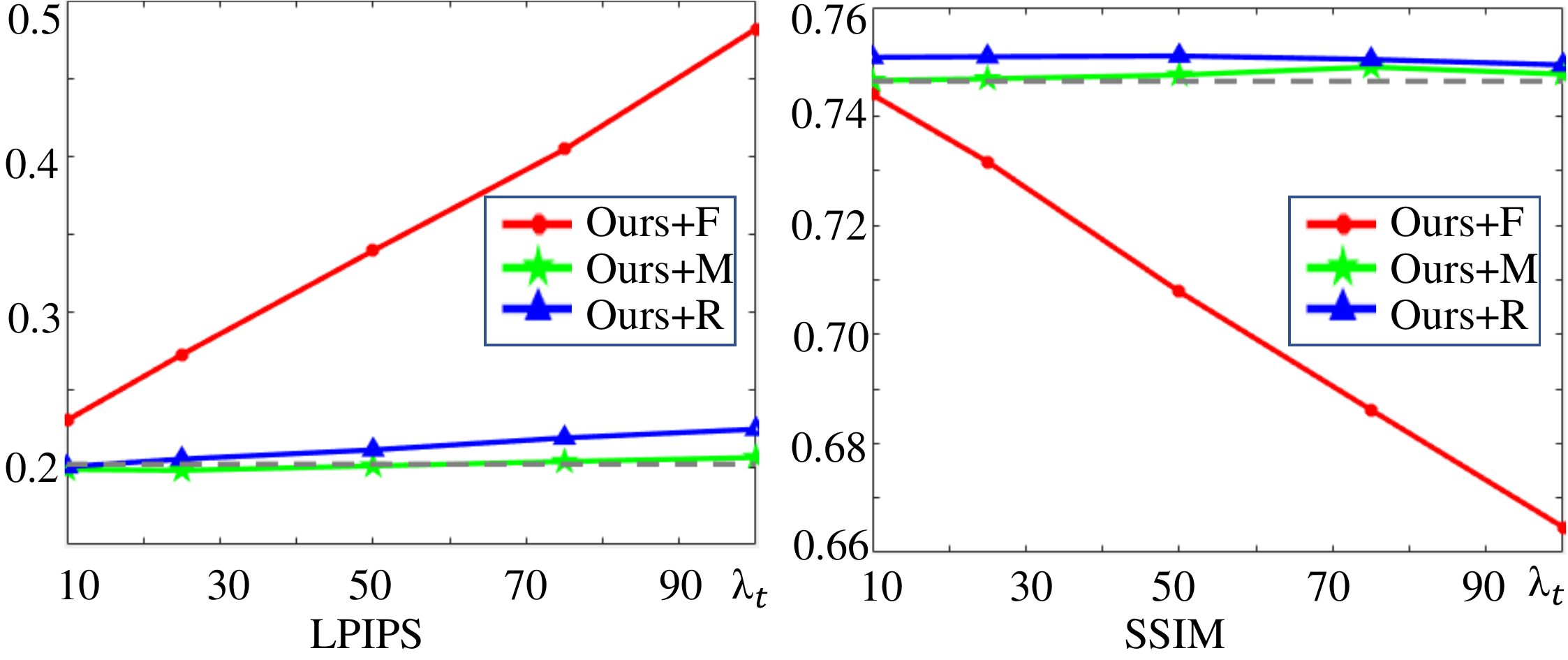}
    \vspace{-0.12in}
    \caption{Demoiréing performance when increasing $\lambda_{t}$.}
    \vspace{-0.22in}
    \label{fig:trend}
\end{figure}

{As shown in Table~\ref{tab: consistency}, the multi-frame demoiréing (Ours) is more consistent than the single-frame demoiréing (Ours\_S). Also, the FID indicates that videos restored by multiple frames are closer to ground-truth videos with higher quality. By incorporating temporal constraints, the video temporal consistency is improved. Specifically, the flow-based method (Ours+F) has the best warping error, but the LPIPS shows that the frame-level quality may drop significantly. Furthermore, only 9\% of users preferred this type of videos when compared with the full version of our method (Ours+M). In contrast, our multi-scale relation-based loss (Ours+M) can improve the video temporal consistency while maintaining the frame-level quality (LPIPS is similar to the method without using temporal consistency regularization, 0.201 v.s. 0.202). More users preferred these results in comparison with all over baselines.} 

\begin{figure}
    \centering
    \includegraphics[width=1.0\linewidth]{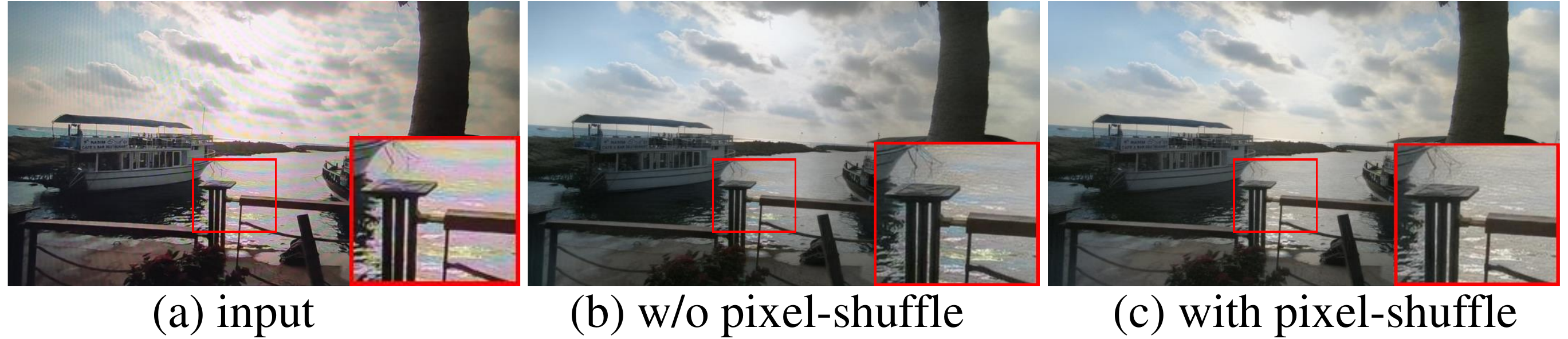}
    \vspace{-0.25in}
    \caption{Different receptive fields. A large receptive field (with pixel-shuffle) benefits the moiré artifacts removal.}
    \vspace{-0.2in}
    \label{fig:shuffle}
\end{figure}

\vspace{-0.15in}
\paragraph{More Analysis on Temporal Consistency.} In the following, we perform more analysis to demonstrate the robustness of our relation-based loss. We plot the curve of demoiréing performance at different weights $\lambda_t$ of the temporal consistency loss. The results are shown in Fig.~\ref{fig:trend}, where the dotted line represents the performance without temporal constraints (Ours).
With the increase of $\lambda_t$, the flow-based (Ours+F) consistency regularization leads to worse LPIPS and SSIM. On the contrary, our multi-scale relation-based approach (Ours+M) learns consistency priors directly from ground-truth videos without sacrificing video quality (please refer to our videos).
 
{We show visual comparisons in Fig.~\ref{fig: relation}.  When compared with reference images (Fig.~\ref{fig: relation} (d)) without temporal constraints (Ours), the flow-based method (Ours+F) heavily blurs image details, such as repetitive textures of the grass and cracks on the stone. By contrast, the multi-scale relation-based method (Ours+M) preserves image details well (Fig.~\ref{fig: relation} (c)), which is comparable to reference images with improved temporal consistency.}
 

\subsection{Ablation Studies}
\label{sec: ablation}
\vspace{-0.05in}
\paragraph{Components of Networks.}
We validate our network designs from the following two aspects. 1) \emph{Receptive field enlargement due to the pixel shuffle operation}: we remove the pixel shuffle operation to reduce the network's receptive field and evaluate the performance.
From results in Table~\ref{tab: ablation}, we observe that the performance degrades without using pixel shuffle. Besides, a large receptive field benefits high-resolution images and large moiré patterns. This can be seen in Fig.~\ref{fig:shuffle}, where moiré artifacts on the lake are removed under the large receptive field.  2) \emph{Analysis of blending weights}: to better understand the role of blending weights in our model, we visualize the weight maps (see Fig.~\ref{fig:weights}) that are used to merge multi-frame features. The weight maps can reflect moiré patterns and choose valuable information from nearby frames for fusion, as shown in Fig.~\ref{fig:weights} (a). Moreover, we compare with a special scenario where the inputs are repetitions of a single frame. 
Under this circumstance, it is difficult to infer moiré patterns without clues from auxiliary frames, as shown in weight maps (Fig.~\ref{fig:weights} (b)). Consequently, the final demoiréing results (Fig.~\ref{fig:weights} last column) become worse.

\vspace{-0.15in}
\paragraph{Deep Supervision Loss.} 
To illustrate this, we build the loss function only on the original image scale. From Table~\ref{tab: ablation}, we observe that the deep supervision loss boosts the performance regarding all three metrics. A possible explanation is that deep supervision loss forces each branch to learn more reasonable demoiréing representations and facilitate the optimization process. 

\begin{table}
    \centering
    \resizebox{.88\columnwidth}{!}{
    \begin{tabular}{c|c|c|c}
    \hline
     Methods & LPIPS $\downarrow$ & PSNR $\uparrow$ & SSIM $\uparrow$\\
     \hline
     no pixel-shuffle & 0.205 & 21.372 & 0.733\\
     no deep supervision loss  & 0.216 & 21.153 & 0.728\\
     \hdashline
     Ours & 0.202 & 21.725 & 0.733\\
     \hline
    \end{tabular}}
    \vspace{-0.1in}
    \caption{Ablation study on the network and loss.}
    \label{tab: ablation}
    \vspace{-0.2in}
\end{table}

\vspace{-0.15in}
\paragraph{Relation-Based Temporal Consistency.}
\label{paragraph: relation}
We validate two variants of relation-based losses: the multi-scale relation-based loss (Ours+M) and the basic relation-based loss (Ours+R). From Fig.~\ref{fig: relation} (b), the textures are a bit blurry with the basic relation-based loss and are worse than results (Fig.~\ref{fig: relation} (c)) from our multi-scale design.  The reason might be that region-level statistics (\ie, mean) help reduce negative impacts of temporal-consistency regularization, which tends to average and erase image details. In comparison with the multi-scale design in Table~\ref{tab: consistency}, fewer users (42\%) selected the basic single-scale design. More importantly, the multi-scale based regularization can well maintain the frame-level qualitative performance (see LPIPS in Fig.~\ref{fig:trend}).




\section{Limitations and Broader Impacts}
Although we have designed a pipeline to ensure the alignment of captured data pairs, it is difficult to perfectly align them under different camera views. Currently, our model also suffers from generalization issues if evaluated on data captured using new devices (\eg, different ISP and Bayer filters) and screens (\eg, different resolution). Expanding the scale of the dataset is one potential solution that will be our future work. In addition, the relation-based loss is generic and can potentially be applied to other video tasks, such as video  stabilization. In practice, we have found that the video instability caused by frame misalignments has been reduced. One possible explanation is that stabilization priors are learned from ground-truth videos.

\section{Conclusion}
In this work, we construct the first video demoiréing benchmark, including a hand-held video demoiréing dataset, and develop a baseline video demoiréing model, effectively leveraging multiple frames. More importantly, we design an effective relation-based consistency regularization, which simultaneously boosts video temporal consistency and maintains visual quality. Detailed analyses are carried out to assist the understanding of video moiré patterns and the weaknesses of flow-based consistency regularization. 
Finally, extensive experiments demonstrate the superiority of our method. 

\vspace{0.05in}
\noindent\textbf{Acknowledgement:} This work is supported by HKU-TCL Joint Research Center for Artificial Intelligence, National Key R\&D Program of China (No.2021YFA1001300), and Guangdong-Hong Kong-Macau Applied Math Center grant 2020B1515310011.


{\small
\bibliographystyle{ieee_fullname}
\bibliography{main}
}

\clearpage
\appendix

\renewcommand{\thesection}{S\arabic{section}}
\renewcommand{\thetable}{S\arabic{table}}
\renewcommand{\thefigure}{S\arabic{figure}}

\noindent\textbf{\Large Outline} 
\vspace{0.1in}
\\In the following, we evaluate our method on another collected dataset in Sec.~\ref{sec: generalization}, incorporate the pre-training into our video demoiréing model in Sec.~\ref{sec: pre-training}, conduct experiments on the low-light video enhancement in Sec.~\ref{sec: low_light}, describe more details about the user studies in Sec.~\ref{sec: Details}, and show more results in Sec.~\ref{sec: more}.

\begin{figure}[!htb]
    \centering
    \includegraphics[width=1.0\linewidth]{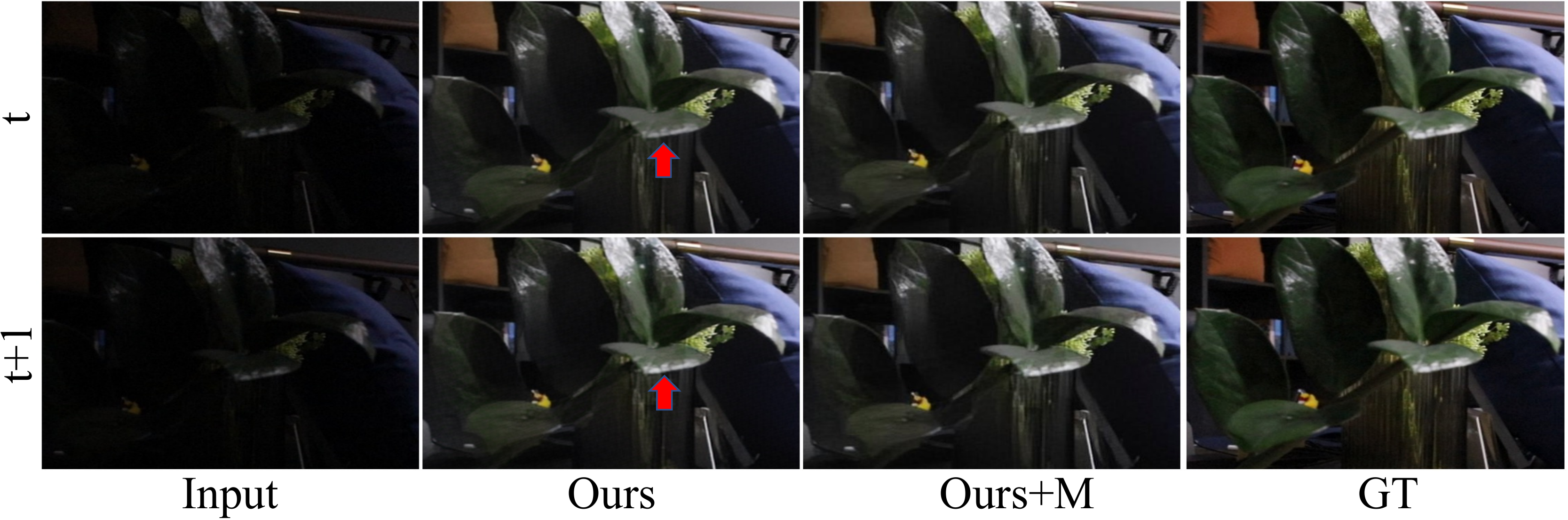}
    \caption{Low-light video enhancement. Our method can also be used to enhance low-light videos and improve the temporal consistency.}
    \vspace{-0.1in}
    \label{fig:low-light}
\end{figure}

\begin{figure*}
    \centering
    \includegraphics[width=1.0\linewidth]{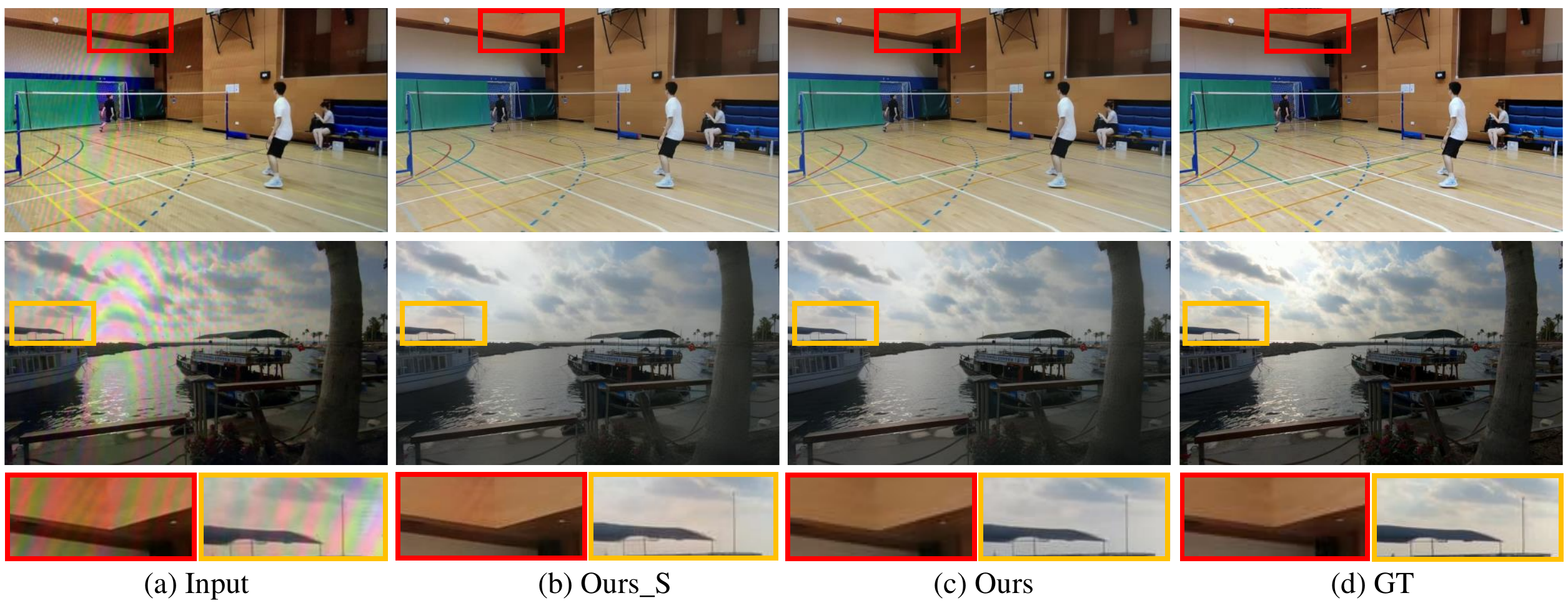}
    \vspace{-0.25in}
    \caption{Demoiréing performance on the iPhone video demoiréing dataset.}
    \vspace{-0.1in}
    \label{fig: iphone_visual}
\end{figure*}

\begin{figure*}
    \centering
    \includegraphics[width=1.0\linewidth]{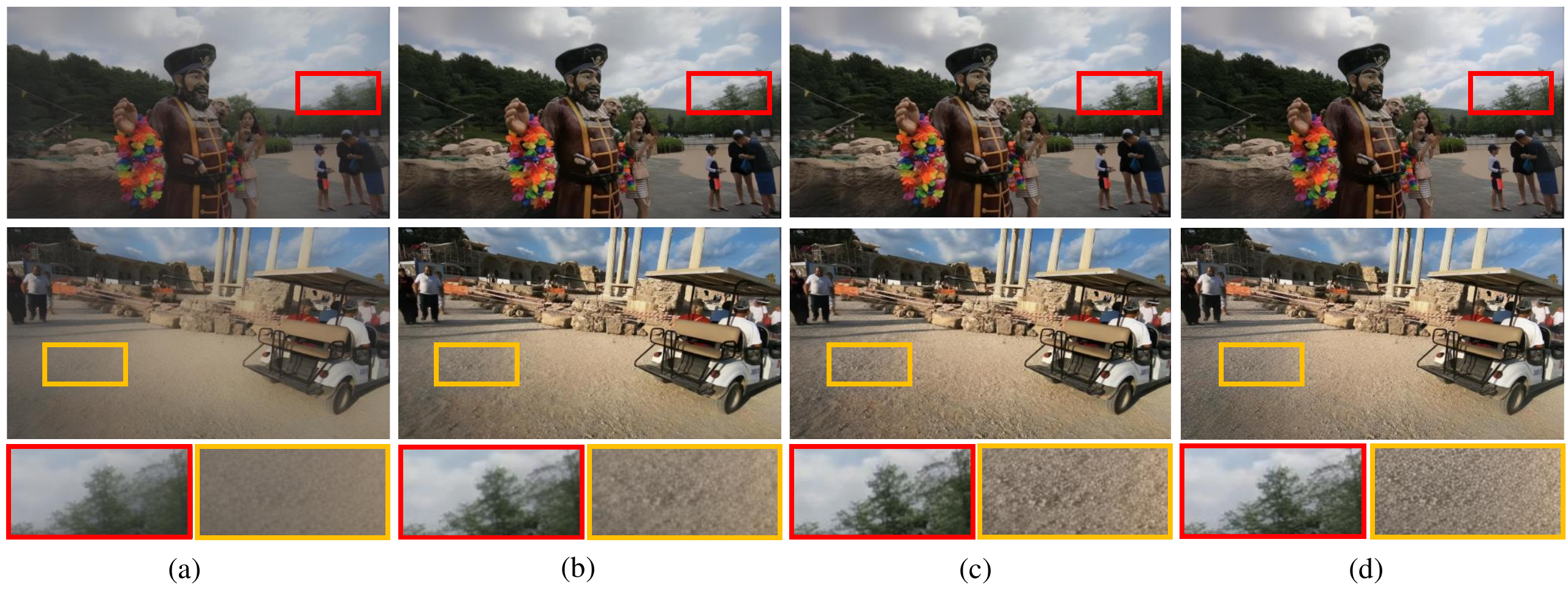}
    \vspace{-0.25in}
    \caption{Different types of temporal consistency on iPhone video demoiréing dataset: (a) flow-based temporal consistency; (b) basic relation-based temporal consistency; (c) multi-scale relation-based temporal consistency; (d) reference without using temporal constraints.}
    \vspace{-0.18in}
    \label{fig: iphone_relation}
\end{figure*}

\section{Evaluation on New Dataset}
\label{sec: generalization}
The results in our main paper are based on the equipment of the Huipu v270 monitor and TCL20 pro mobile-phone.
Here, we evaluate our method on another video demoiréing dataset using the MacBook Pro and iPhoneXR (to be made publicly available). As with the main paper, we conduct several important experiments to validate the frame-level demoiréing performance (Table~\ref{tab: iphone} and Fig.~\ref{fig: iphone_visual}) and video-level temporal consistency (Table~\ref{tab: iphone_consistency} and Fig.~\ref{fig: iphone_relation}).

In Table~\ref{tab: iphone}, our video demoiréing method (Ours) beats the single-frame demoiréing (Ours\_S) on all three metrics, which again proves the superiority of our video demoiréing. Visually, the demoiréd results are cleaner and closer to ground-truth images, as shown in Fig.~\ref{fig: iphone_visual}.

In Table~\ref{tab: iphone_consistency}, our video demoiréing model with multi-scale relation-based loss (Ours+M) obtains the best FID, which indicates higher video quality and temporal consistency. When compared with the baseline without using temporal constraints (Ours), the LPIPS metric shows that only Ours+M can preserve the frame-level quality (0.207 {\vs} 0.206). Qualitatively, image details are better preserved with multi-scale relation-based designs (Fig.~\ref{fig: iphone_relation} (c)) than flow-based (Fig.~\ref{fig: iphone_relation} (a)) and basic relation-based (Fig.~\ref{fig: iphone_relation} (b)) regularization.    

In a nutshell, we maintain similar performance gains as demonstrated in the main paper, which proves the wide applicability of our method. All our data and codes will be publicly available to the community. 

\begin{table}[!htb]
    \centering
    \begin{tabular}{c|c|c|c}
    \hline
     Methods & LPIPS $\downarrow$ & PSNR $\uparrow$ & SSIM $\uparrow$\\
     \hline
     Ours\_S & 0.217 & 22.040 & 0.710\\
     Ours & \textbf{0.206} & \textbf{22.210} & \textbf{0.715}\\
     \hline
    \end{tabular}
    \caption{Demoiréing performance on the iPhone dataset.}
    \label{tab: iphone}
    \vspace{-0.1in}
\end{table}

\begin{table}[!htb]
    \centering
    \setlength{\tabcolsep}{1.2mm}
    \begin{tabular}{c|c|c|c}
    \hline
     Methods & FID $\downarrow$ & warping error $\downarrow$ &  \textcolor[rgb]{0.5,0.5,0.5}{LPIPS$\downarrow$}\\
     \hline
     Ours & 0.091 & 5.26 & \textcolor[rgb]{0.5,0.5,0.5}{0.206}\\
     Ours\_S & 0.099 & 5.80 & \textcolor[rgb]{0.5,0.5,0.5}{0.217}\\
     \hdashline
     Ours+F & 0.110 & \textbf{2.70} &\textcolor[rgb]{0.5,0.5,0.5}{0.328}\\
     Ours+R & 0.089 & 4.40 & \textcolor[rgb]{0.5,0.5,0.5}{0.225}\\
     Ours+M & \textbf{0.088} & 4.70 & \textcolor[rgb]{0.5,0.5,0.5}{0.207}\\
     \hdashline
     GT & 0.000 & 4.56 & \textcolor[rgb]{0.5,0.5,0.5}{0.000} \\
     \hline
    \end{tabular}
    \caption{Temporal consistency on the iPhone dataset ($\lambda_t$: 50).}
    \label{tab: iphone_consistency}
    \vspace{-0.1in}
\end{table}

\section{Pre-training}
\label{sec: pre-training}
Considering that previous works of single-image demoiréing have collected moiré images, such as FHDMi~\cite{he2020fhde}, we investigate whether pre-training on image datasets will further boost the performance. Specifically, we pre-train our model on 9980 moiré images, with each image augmented by random rotation and translation to simulate frame sequences, and then fine-tune the pre-trained weights on our video demoiréing datasets. As shown in Table~\ref{tab: pre-training}, we do not observe significant performance boosts, potentially due to distribution gaps. The question of how to leverage large-scale image demoiréing data to pre-train video demoiréing models is still an open problem and worthy of exploration.

\section{Low-light Video Enhancement.}
\label{sec: low_light}
We conduct experiments on the low-light video enhancement task to demonstrate the generality of the proposed method. Follow the training and testing splits in~\cite{wang2021seeing}, our method successfully enhance low-light video frames (see Fig.~\ref{fig:low-light}). With relation-based loss (Ours+M), flickers are suppressed on leaves and quantitatively reflected by the decreased warping error (2.22 to 2.12). Moreover, FID$\downarrow$ (0.156 to 0.152) is maintained indicating preserved video fidelity.

\begin{table}[!htb]
    \centering
    \begin{tabular}{c|c|c|c}
    \hline
     Methods & LPIPS $\downarrow$ & PSNR $\uparrow$ & SSIM $\uparrow$\\
     \hline
     Ours & \textbf{0.202} & 21.725 & \textbf{0.733}\\
     Ours+pre-training & 0.204 & \textbf{21.759} & 0.732\\
     \hline
    \end{tabular}
    \caption{Demoiréing performance while using pre-training.}
    \label{tab: pre-training}
    \vspace{-0.15in}
\end{table}

\begin{figure}
    \centering
    \includegraphics[width=1.0\linewidth]{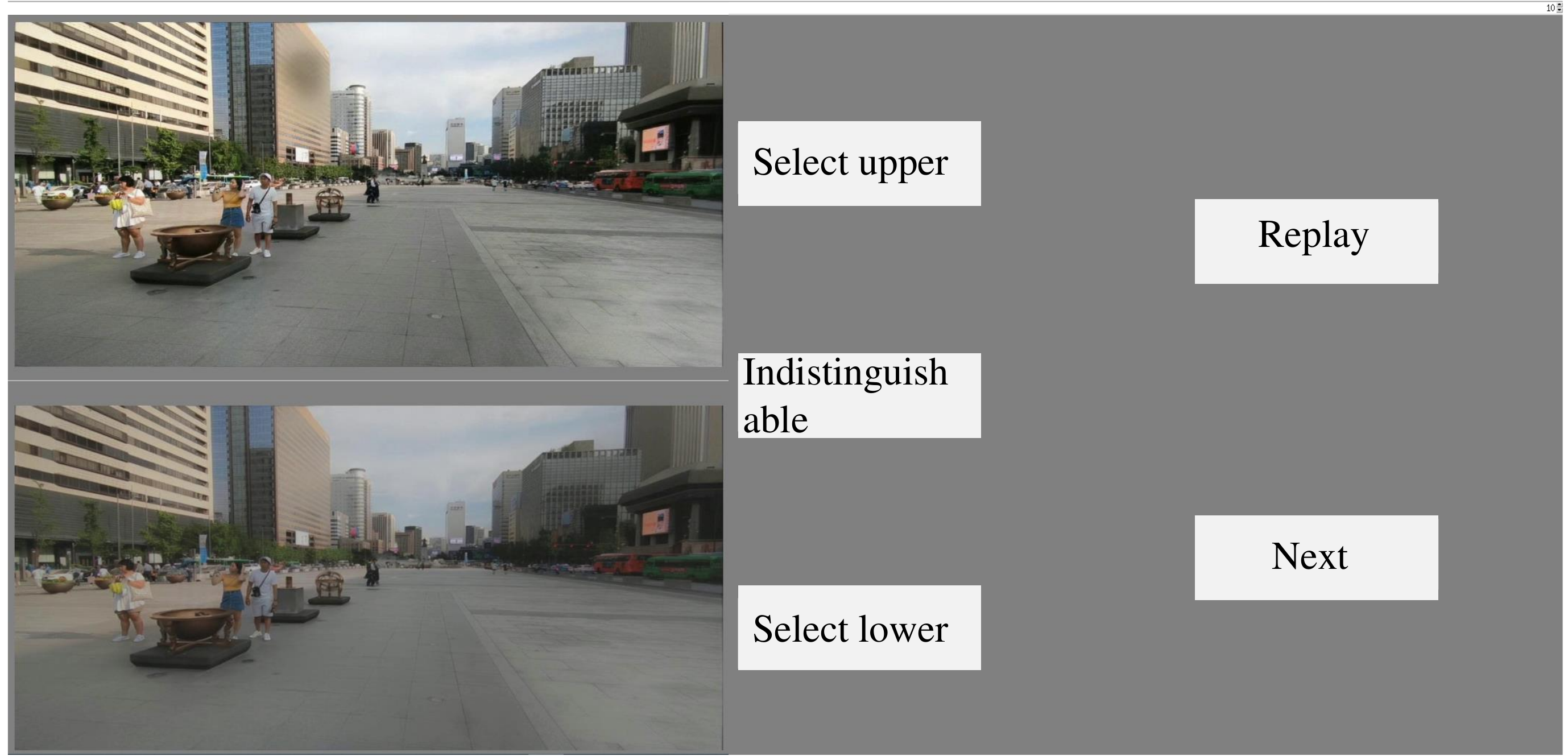}
    \vspace{-0.25in}
    \caption{The interface of user study.}
    \vspace{-0.10in}
    \label{fig: user_study}
\end{figure}

\begin{figure*}
    \centering
    \includegraphics[width=1.0\linewidth]{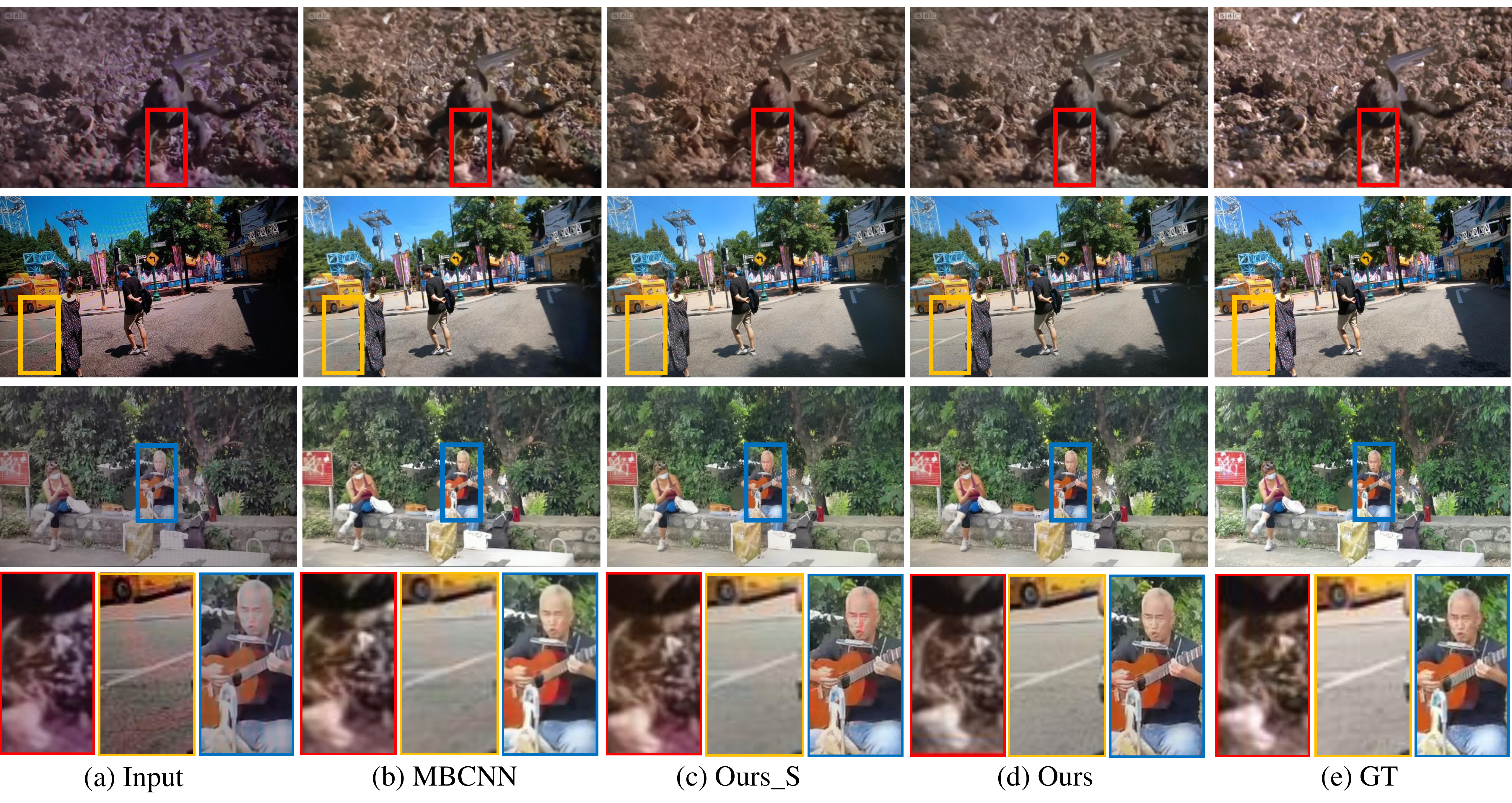}
    \vspace{-0.25in}
    \caption{More Demoiréing results of different methods.}
    \vspace{-0.1in}
    \label{fig: more}
\end{figure*}

\section{Details of User Studies}
\label{sec: Details}
Fig.~\ref{fig: user_study} shows the interface we used for performing user studies. In our experiment, each participant is given 43 video pairs (Ours+M and one of the other methods) for selection. The equipment we used is ASUS ROG ZEPHYRUS, and the frame rate (default 15 fps) and the distance to monitors are not strictly restricted, and participants can move the laptop or adjust the frame rate by clicking on the upper right corner of the interface at any time. If one method is preferred (\ie select upper or select lower), it receives 1 point. 
Otherwise, both methods equally obtain 0.5 points if the 'indistinguishable' button is selected. Finally, we divide the total score by the number of comparisons of one method to get the statistical result. 

\section{More Results}
\label{sec: more}

In Fig.~\ref{fig: more}, we show more results of demoiréd images of different methods. Our method with multiple frames for demoiréing, obtains cleaner results than other single-frame baselines (MBCNN~\cite{zheng2020image} and Ours\_S). More video-level comparisons can be found in our video.

\end{document}